\documentclass[letterpaper]{article} 
\usepackage{aaai24}  
\usepackage{times}  
\usepackage{helvet}  
\usepackage{courier}  
\usepackage[hyphens]{url}  
\usepackage{graphicx} 
\usepackage{natbib}  
\usepackage{caption} 
\usepackage{algorithm}
\usepackage{algorithmic}

\usepackage{tabularx}
\usepackage{makecell}
\usepackage{multirow}
\usepackage{amssymb}
\usepackage{booktabs}
\usepackage{multirow}
\usepackage{tikz}
\usetikzlibrary{positioning, shapes.geometric}
\newcommand{\dataset}{\texttt{FIMO}}

\usepackage{newfloat}
\usepackage{listings}
\DeclareCaptionStyle{ruled}{labelfont=normalfont,labelsep=colon,strut=off} 
\floatstyle{ruled}
\newfloat{listing}{tb}{lst}{}
\floatname{listing}{Listing}
\title{\dataset: A Challenge Formal Dataset for Automated Theorem Proving}
\author {
    Chengwu Liu\textsuperscript{\rm 1},
    Jianhao Shen\textsuperscript{\rm 2},
    Huajian Xin\textsuperscript{\rm 3},
    Zhengying Liu\textsuperscript{\rm 4},
    Ye Yuan\textsuperscript{\rm 1},
    Haiming Wang\textsuperscript{\rm 3},\\
    Wei Ju\textsuperscript{\rm 1},
    Chuanyang Zheng\textsuperscript{\rm 5},
    Yichun Yin\textsuperscript{\rm 4},
    Lin Li\textsuperscript{\rm 2},
    Ming Zhang\textsuperscript{\rm 1}\thanks{Corresponding authors.},
    Qun Liu\textsuperscript{\rm 4}\footnotemark[1],
}
\affiliations {
    \textsuperscript{\rm 1}Peking University,
    \textsuperscript{\rm 2}Huawei HiSilicon,
    \textsuperscript{\rm 3}Sun Yat-sen University,\\
    \textsuperscript{\rm 4}Huawei Noah's Ark Lab,
    \textsuperscript{\rm 5}The Chinese University of Hong Kong\\
    \{liuchengwu, yuanye\_pku, juwei, mzhang\_cs\}@pku.edu.cn,
    \{shenjianhao2, liuzhengying2, yinyichun, lilin29, qun.liu\}@huawei.com,
    \{xinhuajian2000, cyzhengme\}@gmail.com,
    wanghm39@mail2.sysu.edu.cn,
}

\begin{document}

\maketitle

\begin{abstract}

We present \dataset, an innovative dataset comprising formal mathematical problem statements sourced from the International Mathematical Olympiad (IMO) Shortlisted Problems. Designed to facilitate advanced automated theorem proving at the IMO level, \dataset\ is currently tailored for the Lean formal language. It comprises 149 formal problem statements, accompanied by both informal problem descriptions and their corresponding \LaTeX{}-based informal proofs. Through initial experiments involving GPT-4, our findings underscore the existing limitations in current methodologies, indicating a substantial journey ahead before achieving satisfactory IMO-level automated theorem proving outcomes. The dataset has been made available on GitHub\footnote{\url{https://github.com/liuchengwucn/FIMO}} under the terms of the Apache License. 
\end{abstract}

\section{Introduction}

Automated theorem proving is a challenging yet critical field, undergoing rapid evolution and garnering considerable research attention in recent times. While significant progress has been taken since the advent of large language models (LLMs), the challenge of tackling high-school mathematical problems of International Mathematical Olympiad (IMO) level complexity, which demands profound mathematical reasoning and problem-solving capabilities, remains a pivotal open question. Current investigations reveal that LLM-based provers exhibit limited capabilities, managing to address few IMO problems and leaving the majority unexplored. This phenomenon warrants the exploration of several potential factors:

\begin{itemize}
    \item {\bf High cost of formalization.} The realm of formal mathematical data is marked by scarcity since crafting formal mathematical data demands extensive effort from expert humans and consequently comes at a substantial cost. For instance, even one of the most extensive formal mathematics repositories, \textit{mathlib}, a collaborative endeavor aimed at constructing a unified mathematical library in Lean, is only $45$MB in size. In contrast, the GPT-3 training process employs the expansive CommonCrawl dataset\cite{NEURIPS2020_1457c0d6}, which, even after filtration, spans a colossal 570GB, surpassing the size of \textit{mathlib} by over 10,000 times. 
    \item {\bf Low difficulty level of existing datasets.} Existing formal datasets often lean towards lower levels of complexity, potentially hindering models from obtaining the requisite skills. For example, while miniF2F offers $488$ Olympiad mathematical problems, only $40$ of which are from the IMO context. IMO-level challenges inherently pose heightened difficulty, and it is worth noting that few of them have been successfully solved by automatic provers yet~\cite{polu2020generative,zheng2022minif2f,openai-expert,wang-etal-2023-dt}.
    \item {\bf Incompleteness of data.} Most formal mathematical datasets offer formal statements solely, neglecting corresponding statements or proofs in natural languages. This absence not only compromises the statements' interpretability but also hinders neural provers to leverage natural language proofs, a valuable resource often available for high-school Olympiad mathematical problems.
\end{itemize}

IMO stands as a reputable measure of human mathematical proficiency, often serving as an effective predictor of individual capabilities. Inspired by this, we postulate that the mastery of IMO-level intricate mathematical problems could also serve as an indicator of large language models' (LLMs) problem-solving abilities. In response to the aforementioned challenges, we present \dataset, a formalized IMO-level mathematical problem dataset with informal statements and proofs. To lower the cost of formalization, our approach entails the automated formalization (translation from natural language mathematics to formal languages like Lean) of the IMO Shortlisted Problems. These problems are thoughtfully curated by the IMO's problem selection committee, mirroring the complexity of IMO challenges. Leveraging the reflective capabilities inherent in LLMs, we substantially enhance the subset of problems amenable to auto-formalization. The proposed dataset, \dataset, consists of 149 formal statements in Lean as well as the corresponding informal statements and proofs in natural language. Each formal statement undergoes thorough manual examination to ensure a faithful alignment with the original informal version. We conduct baseline experiments with GPT-4 and report the results with case studies to further explore the problems solving ability of current LLMs. The collective findings highlight GPT-4's limited capacity to yield satisfactory results, underlining the enduring challenge of automating theorem proving at an IMO level. 

The contributions of this work are presented as follows:
\begin{itemize}
    \item We present \dataset\ which contains 149 IMO-level challenging formal statements in Lean as well as the corresponding informal statements and proofs in natural language.
    \item We propose an auto-formalization process with a dynamic interplay of human and environmental feedback. The significance of allowing a large language model to refine its outcomes through provided feedback emerges as a pivotal factor influencing the performance of auto-formalization.
    \item We evaluate GPT-4 on our dataset and conduct case studies to analyze the capacity of existing approaches for IMO-level problem-solving. We find that current LLMs struggle to prove IMO-level mathematical statements.
\end{itemize}

\section{Background and Related Work}

\subsection{Formal Mathematics}

As mentioned in \cite{koutsoukou-argyraki2021formalising}, formal mathematics enables computers to verify the correctness of a proof. Formal mathematics can also help mathematicians gain brand new insights even in already familiar topics as well as serve educational purposes. Formal mathematics relies heavily on interactive theorem provers (also referred to as proof assistants). Popular choices include Lean, Isabelle, Coq, and HOL Light. Mathematicians can enter statements and proofs written in a formal domain-specific language (DSL), and the interactive theorem prover could check the correctness of the proofs automatically.

\subsection{Auto-Formalization}

The task of formalization is to turn informal descriptions written in natural languages (human readable \LaTeX{} code) into formally correct and automatically checkable format \cite{10.1007/978-3-030-53518-6_1}. However, there is a vast logical gap between formal language and natural language, since every simple argument should be made explicit in formal language. Therefore, formalization is a tedious task, and performing formalization manually on a large scale is very expensive and time-consuming. Auto-formalization with LLMs proposed in \cite{wu2022autoformalization}, is an attempt to perform formalization utilizing the few-shot learning ability of the large language models by prompting them with several examples of informal and formal statement pairs. In their paper, they dealt with the auto-formalization of theorem statements and claimed that LLMs are able to correctly translate a significant portion ($25.3\%$) of mathematical competition problems perfectly to formal specifications in Isabelle/HOL.

\subsection{Datasets}

\subsubsection{Lean Mathlib}

Mathlib is a community-maintained Lean mathematical library aiming to build a unified library of mathematics formalized in the Lean proof assistant \cite{mathlib2020}. It is the largest collection of mathematics that has been formalized in Lean and contains programming infrastructure, mathematics, and tactics. The IMO Grand Challenge \cite{selsam2020imo} is a proposal that aims to build an AI that can win a gold medal in the IMO competition in a formal-to-formal (F2F) way. Note that as a part of the challenge, there is also a collection of solutions to IMO problems stored in mathlib, containing both the formal provable statements and the formal proofs. At the time of writing, mathlib contains $32$ formal IMO problems with their solutions.

\subsubsection{MATH}

MATH \cite{hendrycksmath2021} is a dataset of $12\,500$ challenging competition mathematics problems, with each problem having a step-by-step solution. These problems are drawn from mathematics competitions including AMC 10, AMC 12, AIME, etc, and they are classified into 5 levels. Both the problem and the step-by-step solution are written in natural language and formatted using \LaTeX{}, without a formal version.

\subsubsection{MiniF2F}

MiniF2F \cite{zheng2022minif2f} is a benchmark of $488$ manually formalized statements of Olympiad-type mathematical competition problems. Each of the statements is formalized by human experts in three different formal languages: Lean, Isabelle, and Coq. MiniF2F draws from Olympiad mathematical problems (AIME, AMC, and IMO) as well as high-school and undergraduate math classes. Despite having $488$ problems in the whole dataset, most of them are relatively trivial to solve, and only $40$ of them are drawn from authentic IMO problems. GPT-f, a fine-tuned version of GPT, is able to solve some problems, but none of the statements extracted from IMO problems is successfully proved, as reported in \cite{zheng2022minif2f}. This emphasizes the challenge of IMO-level mathematical reasoning.

Note that the original version of the miniF2F dataset presented by OpenAI only contains formal statements. A derived version \cite{2210.12283} is also available publicly with the addition of an informal statement and an informal proof for each problem.

\subsection{Neural Theorem Provers}

\subsubsection{Provers with Formal Language}

GPT-f \cite{polu2020generative} is an automated prover for the Metamath formalization language. GPT-f generates original mathematical terms via generation from language models, which is based on GPT-3 and fine-tuned for theorem proving. Polu et al.~\cite{openai-expert} follow GPT-f and use expert iteration to generate more training data and successfully improve the pass rate of language models. DT-solver~\cite{wang-etal-2023-dt} proposes a dynamic tree sampling strategy to guide the search procedure with state confidence and proof-level values.

\subsubsection{Provers with Informal Language}

Draft, Sketch, and Prove (DSP) is a method aimed at using informal proofs as guides for automated theorem proving \cite{2210.12283}. DSP maps informal proofs to formal proof sketches and lets the automated provers focus on easier sub-problems. Instead of having to figure out the whole formal proof from zero to one, DSP takes advantage of the proofs written in natural languages that are already existed. Besides, DSP allows informal proofs either written by humans or generated by a language model, which further extended the application of DSP on problems without existing informal proofs.

\section{\dataset\ Dataset}

\subsection{Dataset Construction}

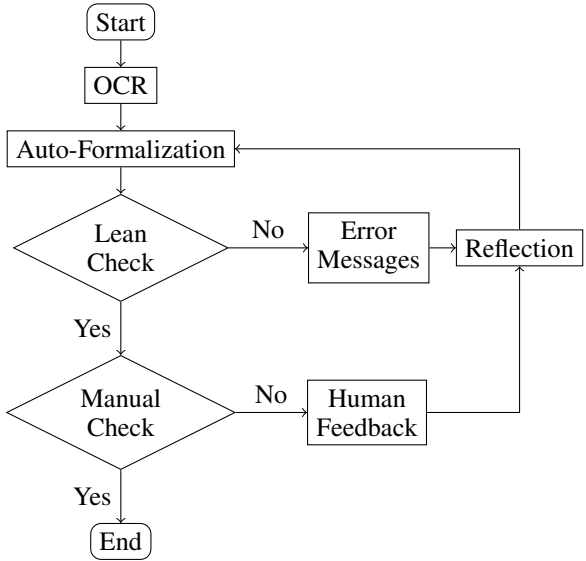
\begin{figure}[h!]
    \centering
    \begin{tikzpicture}[node distance=10pt]
      \node[draw, rounded corners]                                      (start)   {Start};
      \node[draw, below=of start]                                       (ocr)     {OCR};
      \node[draw, below=of ocr]                                         (auto)    {Auto-Formalization};
      \node[draw, align=center, diamond, aspect=2, below=of auto]       (lean)    {Lean\\ Check};
      \node[draw, align=center, right=30pt of lean]                     (env)     {Error\\ Messages};
      \node[draw, right=of env]                           (refine)  {Reflection};
      \node[draw, align=center, diamond, aspect=2, below=20pt of lean]  (manual)  {Manual\\ Check};
      \node[draw, align=center, right=27pt of manual]                   (feedback){Human\\ Feedback};
      \node[draw, rounded corners, below=20pt of manual]                (end)     {End};
      
      \draw[->] (start)  -- (ocr);
      \draw[->] (ocr) -- (auto);
      \draw[->] (auto) -- (lean);
      \draw[->] (lean) -- node[left]  {Yes} (manual);
      \draw[->] (lean) -- node[above] {No}  (env);
      \draw[->] (manual) -- node[left]  {Yes} (end);
      \draw[->] (manual) -- node[above] {No}  (feedback);
      \draw[->] (env) -- (refine);
      \draw[->] (refine) -- (refine|-auto) -> (auto);
      \draw[->] (feedback) -- (feedback-|refine) -> (refine);
    \end{tikzpicture}
    \caption{Flowchart for the pipeline of our auto-formalization with feedback.}
    \label{figure:formalization}
\end{figure}

The process of constructing the dataset can be categorized into three main stages: Optical Character Recognition (OCR), Auto-Formalization with Feedback, and Manual Verification.Each stage is outlined in detail below. An illustrative representation of the entire pipeline is provided in Figure \ref{figure:formalization}.

\subsubsection{OCR}

The IMO Shortlisted Problems are exclusively available in PDF format. To make them amenable to further processing, we employ optical character recognition (OCR) to convert them into \LaTeX{} code. Specifically, the Mathpix snipping tool, an image-to-\LaTeX{} converter, is utilized. This tool seamlessly translates pages of equations into formatted \LaTeX{} code while meticulously retaining all mathematical equation details. In order to avoid minor errors that could potentially compromise mathematical semantics, we subject the generated \LaTeX{} code to manual verification. Following the completion of the aforementioned OCR procedure, we assemble the problems alongside their corresponding solutions. In instances where multiple distinct solutions are provided, our approach is to retain the initial solution while discarding the others. Furthermore, for problems that encompass multiple sub-questions, we opt for simplicity by treating each sub-question as an independent problem. This comprehensive transformation process ensures the conversion of PDF-based IMO Shortlisted Problems into a coherent and structured format, poised for subsequent stages of dataset construction.

The IMO Shortlisted Problems are divided by the IMO problem selection committee into 4 categories: Algebra, Combinatorics, Geometry, and Number Theory. However, our focus for conversion purposes is directed solely toward problems falling under the Algebra and Number Theory domains, along with their corresponding solutions. This selective approach stems from insights outlined in \cite{zheng2022minif2f}, which highlights the formidable challenge of formalization in the Combinatorics and Geometry categories due to the nascent state of formalization efforts in these areas, particularly within formal systems such as Lean. Importantly, it's worth noting that not all Algebra and Number Theory problems inherently necessitate proof-oriented solutions. For instance, a problem may ask students to give some examples or to answer whether a statement is true or not. We follow the method described in \cite{wu2022autoformalization} to address this variability. We apply a transformation to reframe them as proof-oriented problems. Specifically, the solution for each problem is converted into an associated proof. This procedure involves appending the answer to the end of the problem, effectively converting it into a proposition open to formalized proof. An illustrative instance of this conversion process is depicted in Table \ref{table:convert}.

\begin{table}[h!]
    \centering
    \begin{tabularx}{\columnwidth}{c|X}
        \hline
        \multirow{2}{*}{\textbf{\makecell{Original\\ Problem}}} & Find all pairs $(k, n)$ of positive integers for which $7^{k}-3^{n}$ divides $k^{4}+n^{2}$. \\ \hline
        \multirow{3}{*}{\textbf{\makecell{Provable\\ Statement}}} & Find all pairs $(k, n)$ of positive integers for which $7^{k}-3^{n}$ divides $k^{4}+n^{2}$. \textbf{The final answer is $(2,4)$.} \\ \hline
    \end{tabularx}
    \caption{Example of converting a non-proof problem into a provable statement.}
    \label{table:convert}
\end{table}

\subsubsection{Auto-Formalization with Feedback}

Next, our process employs the auto-formalization methodology introduced in \cite{wu2022autoformalization}, leveraging the few-shot learning capabilities inherent in large language models. The few-shot prompts detailed in \cite{wu2022autoformalization} are written in Isabelle notation, which isn't directly compatible with our dataset constructed for the Lean formal language. We manually rewrite the few-shot prompts and replace each Isabelle statement with the corresponding Lean statement. An example of such a rewrite is shown in Table \ref{table:rewrite}. This stage of the dataset construction solidifies the transition from natural language-based problem statements to formalizable propositions, instrumental in furthering our aim of enhancing automated theorem proving capabilities.

\begin{table}[h!]
    \centering
    \begin{tabularx}{\columnwidth}{c|X}
        \hline
        \multirow{2}{*}{\textbf{\makecell{Natural\\ Language\\ Version}}} &  If $3a + b + c = -3, a+3b+c = 9, a+b+3c = 19$, then find $abc$. The final answer is -56 \\ \hline
        \textbf{\makecell{Isabelle\\ Version}} & \makecell[l]{
        \verb|theorem| \\
        \quad\verb|fixes a b c :: real| \\
        \quad\verb|assumes "3*a + b + c = -3"| \\
        \quad\quad\verb|and "a + 3*b + c = 9"| \\
        \quad\quad\verb|and "a + b + 3*c = 19"| \\
        \quad\verb|shows "a * b * c = -56"|
        } \\ \hline
        \textbf{\makecell{Lean\\ Version}} & \makecell[l]{
        \verb|theorem| \\
        \quad\verb|(a b c :|\ \ $\mathbb{R}$\verb|)| \\
        \quad\verb|(|$h_0$\verb| : 3*a + b + c = -3)| \\
        \quad\verb|(|$h_1$\verb| : a + 3*b + c = 9)| \\
        \quad\verb|(|$h_2$\verb| : a + b + 3*c = 19) :| \\
        \quad\verb|a * b * c = -56 :=|
        } \\ \hline
    \end{tabularx}
    \caption{Rewrite few-shot prompts for auto-formalization with Lean language manually.}
    \label{table:rewrite}
\end{table}

Bridging the vast logical gap between natural languages and formal languages is no easy task. Analogous to human comprehension, large language models (LLMs) may initially produce outputs that encompass syntactic or semantic inaccuracies. To ameliorate this issue, we draw inspiration from the concept expounded in \cite{shinn2023reflexion}, wherein verbal reinforcement is employed to facilitate the LLMs' learning from previous inadequacies. The interactive theorem prover Lean could act as the environment and point out syntax errors directly with its error messages. This orchestration enables the direct identification of syntax errors through Lean's error messages, considerably simplifying the process of validating syntactical correctness. Consequently, the focus of human experts can be channeled toward identifying semantic errors, further enhancing the iterative refinement process. We prompt the LLM to refine its results several times, until it is verified by both the proof assistant Lean and human experts or the maximum number of iterations (to be specific, 5) is reached. By this means, we enable the LLMs to refine its result iteratively. An example of such refinement is shown in Table \ref{table:feedback}.

\begin{table}[h!]
    \centering
    \resizebox{\linewidth}{!}{
    \begin{tabularx}{\columnwidth}{c|X}
        \hline
        \multirow{5}{*}{\textbf{\makecell{Informal\\ Statement}}} & Let $X$ be a set of 10000 integers, none of them is divisible by 47 . Prove that there exists a 2007-element subset $Y$ of $X$ such that $a-b+c-d+e$ is not divisible by 47 for any $a, b, c, d, e \in Y$. \\ \hline
        \textbf{\makecell{First\\ Try}} & \makecell[l]{
        theorem exists\_subset\_not\_divisible\_by\_47 : \\
        \textcolor{red}{$\exists$\ (Y : finset\ $\mathbb{N}$)\ (X : finset\ $\mathbb{N}$),} \\
        (finset.card X = 10000)\ $\land$ \\
        ($\forall$ x $\in$ X,\ x \% 47 $\neq$ 0)\ $\land$ \\
        (finset.card Y = 2007)\ $\land$ \\
        (Y $\subseteq$ X)\ $\land$ \\
        ($\forall$ a b c d e $\in$ Y, \\
        (a - b + c - d + e) \% 47 $\neq$ 0) := \\
        } \\ \hline
        \multirow{2}{*}{\textbf{\makecell{Reflective\\ Feedback}}} & \textbf{X is the premise instead of the goal.} Try again. \\ \hline
        \textbf{\makecell{Refined\\ Result}} & \makecell[l]{
        theorem exists\_subset\_not\_divisible\_by\_47 : \\
        \textcolor{red}{$\forall$ (X : finset $\mathbb{N}$),} \\
        (finset.card X = 10000) $\land$ \\
        ($\forall$ x $\in$ X, x \% 47 $\neq$ 0) $\to$ \\
        \textcolor{red}{$\exists$ (Y : finset $\mathbb{N}$),} \\
        (finset.card Y = 2007) $\land$ \\
        (Y $\subseteq$ X) $\land$ \\
        ($\forall$ a b c d e $\in$ Y, \\
        (a - b + c - d + e) \% 47 $\neq$ 0) := \\
        } \\ \hline
    \end{tabularx}}
    \caption{Refine the result with human feedback focusing on semantic errors. The LLMs are able to understand the high-level semantic error described in natural language and to figure out a way to correct it.}
    \label{table:feedback}
\end{table}

Logically, the aforementioned method of auto-formalization with feedback process is LLM-agnostic and any LLM with the ability of few-shot learning should be able to perform the task. Specifically, we use GPT-4 (to be specific, gpt-4-0314) to perform the auto-formalization \cite{openai2023gpt4}.

\subsubsection{Manual Verification}

The training objective of GPT series models focused on maximizing the probability of the next word, which are prone to minor errors. While natural languages are robust to such mistakes, those errors may be extremely harmful to mathematics statements \cite{shen-etal-2021-generate-rank}. Mathematics statements are vulnerable to minor errors, which often change the semantic result in the whole statement being wrong and unprovable. Hence, we introduce multiple manual checks to ensure the formal statements are semantically aligned with the informal ones. Specifically, we manually check the \LaTeX{} code generated by the OCR system (namely the Mathpix snipping tool) and each formal statement generated by the LLM that has passed the verification of the proof assistant Lean, as shown in Figure \ref{figure:formalization}.

\subsection{Statistics}

\dataset\ is a dataset of human-verified auto-formalized statements of IMO-level mathematical problems aligned in Lean language, as well as the natural language version of the problems and solutions. It provides an IMO-level challenge benchmark for mathematical reasoning in both formal language and natural language. We follow the same naming convention as the miniF2F dataset. Each Problem that has been successfully formalized is given a problem name ``\verb|imosl_#year_#category_p#number|". For instance, ``\verb|imosl_2007_number_theory_p6|'' refers to the IMO Shortlisted Problems 2007 Number Theory Problem 6 and ``\verb|imosl_2008_algebra_p3_1|'' refers to the first sub-question of the IMO Shortlisted Problems 2008 Algebra Problem 3.

The IMO Shortlisted Problems are selected by the IMO problems selection committee from problems proposed by each country, namely the long list. The IMO Shortlisted Problems are not released until the following year IMO. As of the time of our work, the IMO Shortlisted Problems with Solutions from 2006 to 2021 are available on the IMO official website\footnote{The IMO Shortlisted Problems with Solutions are available at https://www.imo-official.org/problems.aspx}. Each year, around $30$ problems are selected by the problem selection committee of IMO, and nearly half of them belong to Algebra and Number Theory categories. The total quantity of all shortlisted Algebra and Number Theory problems from 2006 to 2021 is $245$, including $121$ Algebra problems and $124$ Number Theory problems (problems that have multiple sub-questions are treated as multiple independent problems). With auto-formalization with feedback described above, we successfully formalize over $60\%$ ($149/245$) of the problems. The statistics of the proposed \dataset\ dataset and the comparison with other datasets are shown in Table \ref{table:stat} and Table \ref{table:compare}.

\begin{table}[h!]
    \centering
    \begin{tabular}{cccc}
    \toprule
    \textbf{Category} & \textbf{\makecell{Total\\ Quantity}} & \textbf{\makecell{Success\\ Count}} & \textbf{\makecell{Success\\ Rate}} \\
    \midrule
    {\textbf{Algebra}} & {$124$} & {$89$} & {$71.8\%$} \\ 
    \textbf{\makecell{Number Theory}} & $121$ & $60$ & $49.6\%$ \\
    \midrule
    \textbf{Total} & $245$ & $149$ & $60.8\%$ \\
    \bottomrule
    \end{tabular}
    \caption{The success rate of GPT-4 auto-formalization with reflective feedback. Success count refers to the number of problems that are successfully formalized. Since problems that have multiple sub-questions are treated as multiple independent problems, the total quantities of problems are slightly larger.}
    \label{table:stat}
\end{table}

\begin{table}[h!]
    \centering
    \resizebox{\linewidth}{!}{
    \begin{tabular}{ccccccc}
    \toprule
    \textbf{Dataset} & \textbf{\makecell{Problem\\ Count}} & \textbf{\makecell{IMO-\\ Level}} & \textbf{FS} & \textbf{FP} & \textbf{IS} & \textbf{IP} \\
    \midrule
    \textbf{mathlib} & N/A & $32$ & $\checkmark$ & $\checkmark$ & -- & -- \\
    \textbf{miniF2F} & $488$ & $40$ & $\checkmark$ & -- & --* & --* \\
    \textbf{MATH} & $12\,500$ & $0$ & -- & -- & $\checkmark$ & $\checkmark$ \\
    \midrule
    \textbf{\dataset} & $149$ & $149$ & $\checkmark$ & -- & $\checkmark$ & $\checkmark$ \\
    \bottomrule
    \end{tabular}}
    \caption{Comparison with other datasets. FS, FP, IS and IP refer to formal statement, formal proof, informal statement, and informal proof, respectively. IMO-Level refers to problems that are either authentic IMO problems or IMO Shortlisted Problems which are both selected by the problem selection committee of the IMO. The asterisks mean that there is a derived version that contains informal statements and informal proofs, while the original version does not.}
    \label{table:compare}
\end{table}

\subsection{Dataset Analysis}

Translating mathematical statements into formal language manually is a tedious and time-consuming process. With the few-shot learning ability of LLMs, we can prompt LLMs with some examples and auto-formalize the statements. However, only a little proportion ($25.3\%$) of the statements could be formalized successfully as reported in \cite{wu2022autoformalization}. To address this problem, we prompt the LLMs with the feedback that comes from either the error messages generated by the proof assistant Lean or written by a human expert, as shown in Figure \ref{figure:formalization}. 

We also report the success rate of the auto-formalization without reflective feedback. The result shows that there will be a significant drop in the success rate without feedback, demonstrating the effectiveness of our method. The result also shows that Algebra problems are generally easier to formalize than Number Theory problems. The success rate of the auto-formalization process of Algebra problems also gains a greater boost with reflective feedback. With our method, GPT-4 is able to formalize a considerably larger proportion of problems (from $32.6\%$ to $60.8\%$) while keeping the need for human intervention comparable.

\begin{table}[h!]
    \centering
     \resizebox{\linewidth}{!}{
     \begin{tabular}{cccc}
    \toprule
    \textbf{Category} & \textbf{\makecell{Total\\ Quantity}} & \textbf{\makecell{Success\\ Count}} & \textbf{\makecell{Success\\ Rate}} \\
    \midrule
    {\textbf{Algebra}} & {$124$} & {$52$ ($-37$)} &{$41.9\%$ ($-29.9\%$)} \\ 
    \textbf{\makecell{Number Theory}} & $121$ & $28$ ($-32$) & $23.1\%$ ($-26.5\%$) \\
    \midrule
    \textbf{Total} & $245$ & $80$ ($-69$) & $32.6\%$ ($-28.2\%$) \\
    \bottomrule
    \end{tabular}}
    \caption{The success rate of GPT-4 auto-formalization without reflective feedback.}
    \label{table:nofeedback}
\end{table}

In order to understand the reason why LLMs fail to formalize mathematical statements, we randomly sample 50 auto-formalized statements that fail to pass the Lean Check. The Lean Proof assistant will point out the syntax error in Lean language, and generate corresponding error messages. The syntax error can be divided into 7 categories, as shown in Table \ref{table:errortype}.

\begin{table}[h!]
    \centering
    \resizebox{\linewidth}{!}{
    \begin{tabularx}{\columnwidth}{Xccc}
    \toprule
    \textbf{Error Type} & \textbf{Algebra} & \textbf{\makecell{Number\\ Theory}} & \textbf{Total} \\ \midrule
    failed to synthesize type class instance & \multirow{2}{*}{6} & \multirow{2}{*}{1} & \multirow{2}{*}{7} \\
    \multirow{2}{*}{\textbf{type mismatch}} & \multirow{2}{*}{6} & \multirow{2}{*}{10} & \multirow{2}{*}{16}\\ \\
    \multirow{2}{*}{\textbf{unknown identifier}} & \multirow{2}{*}{9} & \multirow{2}{*}{8} & \multirow{2}{*}{17}\\ \\
    \multirow{2}{*}{invalid field notation} & \multirow{2}{*}{1} & \multirow{2}{*}{0} & \multirow{2}{*}{1}\\ \\
    \multirow{2}{*}{function expected} & \multirow{2}{*}{1} & \multirow{2}{*}{1} & \multirow{2}{*}{2}\\ \\
    \multirow{2}{*}{unexpected token} & \multirow{2}{*}{2} & \multirow{2}{*}{0} & \multirow{2}{*}{2}\\ \\
    don't know how to synthesize placeholder & \multirow{2}{*}{0} & \multirow{2}{*}{4} & \multirow{2}{*}{4} \\
    \multirow{2}{*}{invalid expression} & \multirow{2}{*}{0} & \multirow{2}{*}{1} & \multirow{2}{*}{1}\\ \\ \midrule
    \multirow{2}{*}{\textbf{Total}} & \multirow{2}{*}{25} & \multirow{2}{*}{25} & \multirow{2}{*}{50}\\ \\
    \bottomrule
    \end{tabularx}}
    \caption{Statistics of the syntax error type extracted from the error messages generated by proof assistant Lean.}
    \label{table:errortype}
\end{table}

As shown in Table \ref{table:errortype}, the most common syntax errors are ``type mismatch'' and ``unknown identifier''. These errors may be due to the hallucination inside the large language model. However, like compilers for other programming languages, the error messages are not always very useful. For instance, an unknown identifier may be indirectly caused by not importing a certain library, or by not declaring a variable properly. We found that GPT-4 is able to figure out how to fix syntax errors under the guidance of high-level error messages in most cases.

We also found that instead of simply mapping natural mathematics into formal mathematics, GPT-4 is able to extract nontrivial latent semantic information. As shown in Table \ref{table:latent}, the informal mathematics statement ``Let $a, b, c$ be the sides of a triangle.'' indicates that $a, b, c$ are all positive real numbers. GPT-4 is aware of it and also notices the fact that the sum of any two sides is greater than the third side. Such auto-formalization requires the model to have both mathematical reasoning ability and general knowledge of mathematics.

\begin{table}[h!]
    \centering
    \begin{tabularx}{\columnwidth}{c|X}
        \hline
        \multirow{8.5}{*}{\textbf{\makecell{Informal\\ Statement}}} & Let $a, b, c$ be the sides of a triangle. Prove that
$$
\frac{\sqrt{b+c-a}}{\sqrt{b}+\sqrt{c}-\sqrt{a}}+\frac{\sqrt{c+a-b}}{\sqrt{c}+\sqrt{a}-\sqrt{b}}
$$
$$
+\frac{\sqrt{a+b-c}}{\sqrt{a}+\sqrt{b}-\sqrt{c}} \leq 3
$$\\ \hline
        \textbf{\makecell{Formal\\ Statement}} & \makecell[l]{
        \verb|theorem imosl_2006_algebra_p5| \\
        \quad\verb|(a b c : |$\ \mathbb{R}$\verb|)| \\
        \textcolor{red}{\quad\verb|(|$h_0$\verb| : a > 0 |\ $\land$\verb| b > 0 |\ $\land$\verb| c > 0)|} \\
        \textcolor{red}{\quad\verb|(|$h_1$\verb| : a + b > c |\ $\land$\verb| a + c > b|} \\
        \textcolor{red}{\quad$\land$\verb| b + c > a) :|} \\
        \quad\verb|(real.sqrt (b + c - a) / | \\
        \quad\verb|(real.sqrt b + real.sqrt c | \\
        \quad\verb|- real.sqrt a)) +| \\
        \quad\verb|(real.sqrt (c + a - b) / | \\
        \quad\verb|(real.sqrt c + real.sqrt a | \\
        \quad\verb|- real.sqrt b)) +| \\
        \quad\verb|(real.sqrt (a + b - c) / | \\
        \quad\verb|(real.sqrt a + real.sqrt b | \\
        \quad\verb|- real.sqrt c)) ≤ 3 :=| \\
        \verb|begin| \\
        \quad\verb|sorry| \\
        \verb|end| \\
        } \\ \hline
    \end{tabularx}
    \caption{Instead of simply mapping natural mathematics into formal mathematics, LLMs can extract latent semantic information. The original problem is the IMO Shortlisted Problems 2006 Algebra Problem 5.}
    \label{table:latent}
\end{table}

\section{Experiments and Discussion}\label{analysis}

In this section, in order to understand the performance of existing LLMs, we conducted baseline experiments using the cutting-edge GPT-4 (to be specific, gpt-4-0314) as our baseline model. Current methods of automated theorem proving with LLMs can be divided into two classes in accordance with the usage of data. The first class of methods only uses formal data in the process of the stage of model fine-tuning and inference. 
Well-known methods include PACT \cite{han2022proof} and Thor \cite{51693}. PACT extracts self-supervised data from formal proof terms for training. Thor leverages the power of automated theorem provers instead of language models for premise selection.
On the other hand, the second class of methods leverages both formal data and informal data and has potential to achieve better performance.
These methods include GPT-f and DSP. GPT-f uses the WebMath dataset, a mixture of formal data and informal data, in their pre-training process. The source of data involved in the WebMath dataset includes GitHub, arXiv Math, and Math StackExchange. In the DSP, the formal proofs are generated under the guidance of the existing informal proofs.
Since the \dataset\ dataset proposed contains both formal data and informal data, it supports both classes of methods.
Inspired by the DSP framework, we also leverage the power of the existing informal proofs. 
The original version of the DSP framework exclusively targets the formal language Isabelle with Sledgehammer, a proof automation tool in Isabelle. Because it relies on Isabelle instead of Lean, we cannot directly apply it to our dataset.
Our methods are described below.
We conducted error analyses of formal proofs generated by GPT-4 and analyzed possible reasons for the failure of the IMO-level automated theorem proving. We also analyzed the limitation of the \dataset\ dataset.

\subsection{GPT-4 Guided by Informal Proofs}

To guide the LLMs with informal proofs, our approach can be divided into two phases: acquiring the informal proof and formal proof generation.

The first phase is to find informal proofs for a given problem. 
In common scenarios, solutions to the Olympiad mathematical problems are available, as in our dataset. 
But sometimes they can be missing or not suitable for formal expression. Therefore, we conduct experiments in two different setups: with or without human informal proof.
In the former setup, the informal proofs are written by a human. In the \dataset\ dataset, the informal proofs are extracted from the solutions provided by the problem selection committee of IMO.
We treat the solution provided as a ``ground-truth'' informal proof. Hopefully, the manually-written informal proof would act as a guide for the process of formal proof generation.
In the latter setup, the informal proofs are generated directly by a language model. We also use GPT-4 for the generation of informal proofs.
We regenerate the informal proof every time we generate the formal proof.

The second phase of our approach is to write formally verifiable proofs. We prompt the language model with formal and informal statements, as well as informal proof acquired in the former phase. In such a way, the language model could obtain the proof ideas from the informal proofs and use them to guide the process of formal proof generation. Our approach enjoys both the logical rigor provided by formal systems and the flexibility of informal proofs. The experiment's result shows that our approach can achieve competitive results on the miniF2F benchmark, without sophisticated search algorithms or further fine-tuning of LLMs.

\subsection{Result}

We report the pass rate of GPT-4 guided by informal proofs on our dataset and both the test set and the validation set of the benchmark miniF2F.
The results are reported in Table \ref{table:result}. Pass rates are reported as Pass@N where N is the number of attempts for generating the whole proof.

GPT-4 achieved a Pass@8 of over 20\% on miniF2F. Nonetheless, GPT-4 failed to prove any of the statements in the \dataset\ dataset as expected, showing the difficulty of the IMO-level problem theorem proving. We also find that the GPT-4 without using human informal proofs achieves better Pass@8. This phenomenon can be attributed to the increased diversity and ability to explore various directions exhibited in informal proofs generated by language models.

\begin{table*}[h!]
    \centering
    \begin{tabular}{ccccccccc}
    \toprule
    \multirow{2}{*}{\textbf{Method}} & \multicolumn{3}{c}{\textbf{miniF2F-valid}} & \multicolumn{3}{c}{\textbf{miniF2F-test}}  & \multicolumn{2}{c}{\textbf{\dataset}}  \\
    & Pass@1 & Pass@8 & Length & Pass@1 & Pass@8 & Length & Pass@1 & Pass@8\\
    \midrule 
    \textbf{\makecell{GPT-4 w/ Human Proof}} & 11.5\% & 16.0\% & 11.1 & 8.6\% & 14.3\% & 11.8 & 0.0\% & 0.0\% \\
    \textbf{\makecell{GPT-4 w/o Human Proof}} & 9.4\% & 21.3\% & 11.2 & 9.0\% & 20.9\% & 11.4 & 0.0\% & 0.0\% \\
    \bottomrule
    \end{tabular}
    \caption{Baseline performance of GPT-4 with/without human informal proofs. Pass rates are reported as Pass@N, where N is the number of attempts for generating proofs. The length refers to the average number of lines of the generated proof.}
    \label{table:result}
\end{table*}

The results show that while GPT-4 is able to solve some problems in the miniF2F dataset, it cannot solve any problem in the \dataset\ dataset.
We believe that this is because statements in our dataset are more difficult to prove and require advanced mathematical reasoning skills. In the following section, we analyze the reason why GPT-4 fails in detail.

\subsection{Error Analysis}

We found that most of the informal proofs synthesized by the language model are mathematically incorrect. Therefore, the LLMs may be misled when generating formal proofs.

\begin{figure}[h!]
    \centering
    \includegraphics[width=\columnwidth]{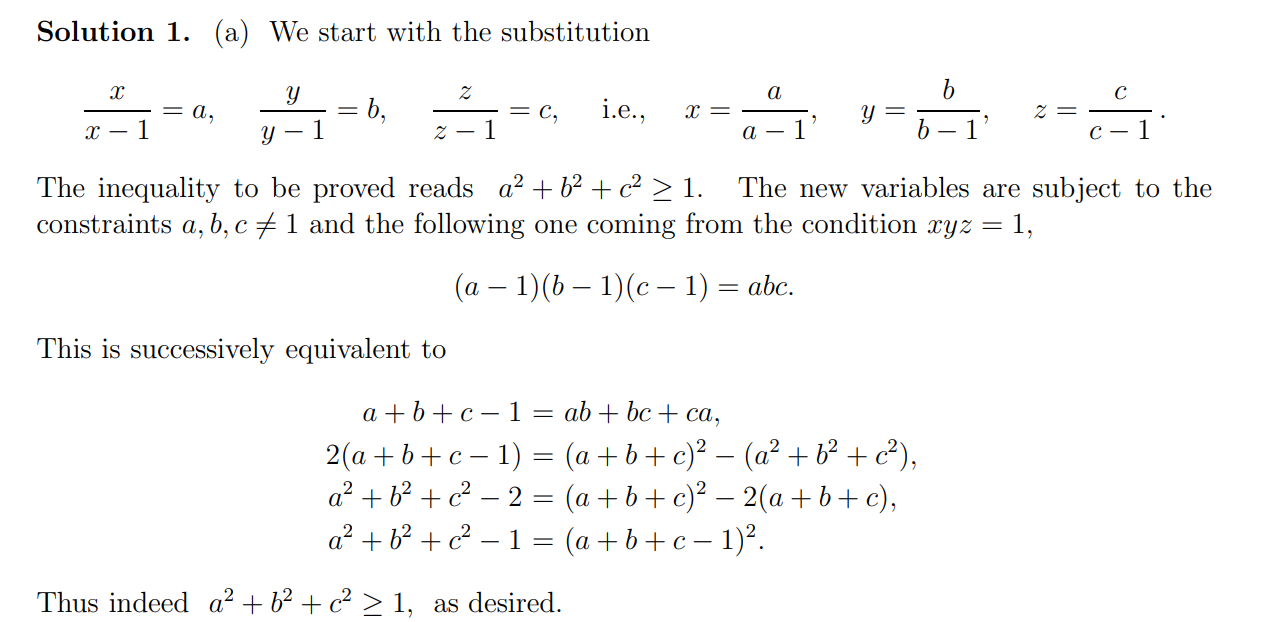}
    \includegraphics[width=\columnwidth]{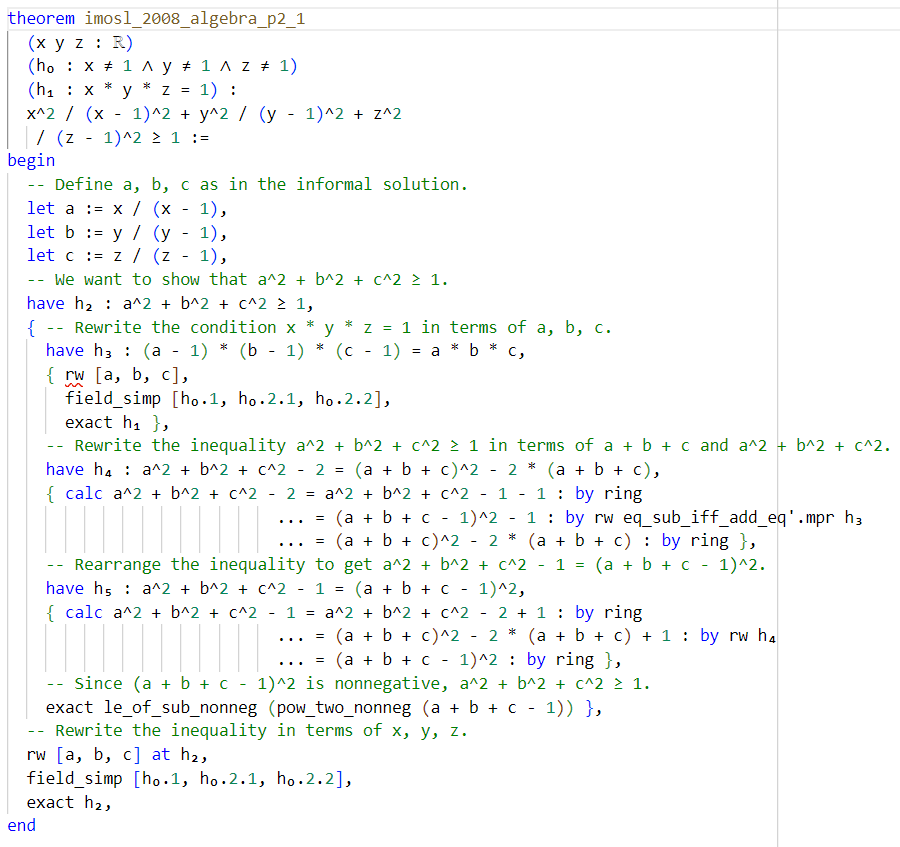}
    \caption{Informal solution provided and formal proof generated. The original problem is the first sub-question of the IMO Shortlisted Problems 2008 Algebra Problem 2.}
    \label{figure:example}
\end{figure}

For GPT-4 with the guidance of human proofs, we observed that the language model is able to follow the solution provided, but fails to apply tactics properly. Figure \ref{figure:example} shows an example emphasizing this kind of failure. The model managed to divide the whole problem into several sub-goals according to the solution provided and explain the motivation of each step in the comments. For instance, the formal proof generated defines $a, b, c$ as in the informal solution, and tries to show $a^2 + b^2 + c^2 \ge 1$. Unfortunately, the formal proof fails to apply the tactic \verb|rw| properly, as indicated by the red underline in the example. We also observed that although there are some errors, GPT-4 is able to generate much longer formal proofs (with an average length of over 11) with the guidance of informal proofs. In contrast, the average proof length is less than 3 for GPT-f on the miniF2F dataset as reported in \cite{zheng2022minif2f}.

\section{Limitation}

Like manually formalized miniF2F, Combinatorics and Geometry problems are not covered by the proposed \dataset\ dataset since they are less expressible in formal systems like Lean. The shift of the distribution of problem type may affect the evaluation of the reasoning ability of the models when the \dataset\ dataset serves as a metric. The dataset covers the IMO Shortlisted Problems with Solutions only from 2006 to 2021, which are publicly available on the official IMO website. At the time of writing, problems before 2006 that can be acquired through other channels are not included. All of the formal statements are formalized automatically and checked by humans. Despite our best efforts, there may still be a very small number of errors.

\section{Conclusion}

We presented \dataset, a dataset of formal IMO-level mathematics problem statements, alongside informal statements and informal proofs in \LaTeX{}. This effort aims to enhance neural mathematical reasoning by bridging natural and formal languages. Through auto-formalization with feedback, we highlighted the significance of reflective feedback in improving the auto-formalization process. Employing GPT-4 guided by informal proofs as our baseline model, we find that GPT-4 falls short of formally proving any statements in the dataset. This outcome emphasizes the enduring challenge of achieving IMO-level automated theorem proving. Our contribution, \dataset, along with our methodology and findings, enriches the landscape of neural mathematical reasoning. While current capabilities have limitations, they underscore the need for continued innovation and research to advance the realm of AI-powered mathematical formalization, thereby narrowing the gap between human expertise and machine capabilities.

\bibliography{aaai24}

\end{document}